\pdfoutput=1

\documentclass[11pt]{article}

\usepackage[preprint]{acl}

\usepackage{tipa}
\usepackage{amsmath,amssymb}
\usepackage{booktabs}     
\usepackage{multirow}     
\usepackage{array}        

\usepackage{times}
\usepackage{latexsym}
\usepackage{hyperref}
\usepackage[T1]{fontenc}

\usepackage[utf8]{inputenc}

\usepackage{microtype}

\usepackage{inconsolata}

\usepackage{graphicx}

\title{Breaking the Transcription Bottleneck: Fine-tuning ASR Models for Extremely Low-Resource Fieldwork Languages}

\author{Siyu Liang \and Gina-Anne Levow \\
University of Washington \\
\texttt{liangsy, levow@uw.edu} \\}

\begin{document}
\maketitle
\begin{abstract}
Automatic Speech Recognition (ASR) has reached impressive accuracy for high-resource languages, yet its utility in linguistic fieldwork remains limited. Recordings collected in fieldwork contexts present unique challenges, including spontaneous speech, environmental noise, and severely constrained datasets from under-documented languages. In this paper, we benchmark the performance of two fine-tuned multilingual ASR models, MMS and XLS-R, on five typologically diverse low-resource languages with control of training data duration. Our findings show that MMS is best suited when extremely small amounts of training data are available, whereas XLS-R shows parity performance once training data exceed one hour. We provide linguistically grounded analysis for further provide insights towards practical guidelines for field linguists, highlighting reproducible ASR adaptation approaches to mitigate the transcription bottleneck in language documentation.
\end{abstract}

\section{Introduction}

Automatic Speech Recognition (ASR) has achieved significant breakthroughs in recent years, with deep learning-based models reported to reach near-human word error rates for high-resource languages \citep{radford_robust_2023, baevski_wav2vec_2020}. However, these advancements have largely been driven by massive transcribed datasets (e.g. \citet{chang_end--end_2022, panayotov_librispeech_2015, godfrey_switchboard_1992}), leaving a substantial performance gap for low-resource languages, particularly those encountered in linguistic fieldwork \citep{guillaume_plugging_2022}. Fieldwork speech data presents distinct challenges, including spontaneous speech, varied recording setups, and typologically diverse linguistic features, all of which could degrade the performance of ASR models trained on standardized speech corpora.

Linguistic fieldwork plays a critical role in preserving endangered languages and documenting linguistic diversity. These recordings capture not only the linguistic structures of a language, but also oral traditions, discourse patterns, and sociolinguistic variations \citep{himmelmann_documentary_1998, austin_cambridge_2011}. However, while some well-researched low-resource languages have substantial datasets \citep{guillaume_fine-tuning_2022, przezdziak_optimizing_2024}, there is usually limited data to bootstrap an ASR model for most field linguists. Evaluations of the ASR approaches usually tend to focus on one language \citep{jones_comparing_2024, rijal_whisper_2024, guillaume_fine-tuning_2022, mainzinger_fine-tuning_2024} or are inconsistent regarding data size, genre, etc. in the sample \citep{jimerson_unhelpful_2023}. Evaluations of models for low-resource languages also tend to favor clean, good quality, read speech  \citep{rijal_whisper_2024, mainzinger_fine-tuning_2024, jimerson_unhelpful_2023}, compared with the noisier and more spontaneous speech of fieldwork recordings.

\subsection{The transcription bottleneck}
Linguistic fieldwork plays a crucial role in documenting endangered and under-researched languages, yet the process of manually transcribing recordings remains a significant barrier: transcribing a single hour of audio in a newly documented language can require up to 50 hours of work \citep{shi_leveraging_2021}. Moreover, many of the fieldwork languages also lack standardized orthographies, requiring a handful of trained linguists to make discerning decisions during transcription. As a result, the volume of untranscribed linguistic data continues to grow, creating a severe bottleneck in language documentation, analysis, and distribution \citep{anastasopoulos_leveraging_2018, bird_sparse_2020, thieberger_oxford_2012}.

The dependence on large transcribed datasets for training ASR models exacerbates this issue, as most endangered and low-resource languages lack sufficient annotated speech data to support model development \citep{levow_developing_2021}. Without adequate transcriptions, traditional supervised ASR methods remain ineffective, requiring alternative approaches that can leverage limited data more efficiently \citep{dunbar_zero_2019, baevski_wav2vec_2020, baevski_unsupervised_2021}.

\subsection{ASR for Low-resource Data}
The application of ASR in linguistic fieldwork closely parallels the development of low-resource ASR research. Since the 2010s, much of this work has focused on languages from the IARPA Babel project, which served as a cornerstone for ASR development in low-resource settings \citep{miao_deep_2013, cui_improving_2014, grezl_adaptation_2014}. Research leveraging Babel datasets introduced key techniques such as transfer learning, multilingual adaptation, and data augmentation, which have since become fundamental to ASR advancements in under-documented languages \citep{zhang_improving_2014,khare_low_2021, vanderreydt_transfer_2022, guillaume_plugging_2022}.

The widespread adoption of the Kaldi toolkit \citep{povey_kaldi_2011} further propelled ASR research in these domains, enabling the development of reproducible pipelines and fostering the open distribution of Kaldi-compatible datasets \citep{milde_open_2018, yadava_development_2017, zhang_wenetspeech_2022}. Concurrently, researchers have explored approaches such as transfer learning and fine-tuning from multilingual pre-trained models \citep{guillaume_plugging_2022, sikasote_bembaspeech_2021} or adapting English-centric models to new linguistic domains \citep{kim_semi-supervised_2021, thai_fully_2020}. Additionally, self-supervised and semi-supervised learning approaches have gained traction as viable solutions for overcoming transcription scarcity, further bridging the gap between ASR and field linguistics \citep{babu_xls-r_2021, baevski_unsupervised_2021}.

\subsection{Fine-tuning Pre-trained ASR Models}

Fine-tuning pre-trained ASR models has emerged as a key approach for improving recognition accuracy in low-resource settings, particularly for linguistic fieldwork recordings \citep{guillaume_plugging_2022, pillai_multistage_2024, nowakowski_adapting_2023}. Self-supervised learning models, such as wav2vec 2.0 \citep{baevski_wav2vec_2020}, HuBERT \citep{hsu_hubert_2021}, MMS (Massive Multilingual Speech) Model \citep{pratap_scaling_2024}, and XLS-R \citep{babu_xls-r_2021}, have demonstrated the ability to learn generalized speech representations from large-scale multilingual datasets, significantly reducing the need for extensive transcriptions in under-documented languages. Studies on specific low-resource languages, such as Bribri \citep{coto-solano_explicit_2021}, Japhug \citep{guillaume_fine-tuning_2022}, Mvskoke \citep{mainzinger_fine-tuning_2024}, and the Čakavian dialect of Croatian \citep{jones_comparing_2024}, underscore the benefits of adapting large multilingual models and report considerable reductions in error rates even with very limited data.

Nevertheless, recent work suggests that no single architecture or end-to-end approach consistently outperforms others under extremely low-resource conditions \citep{jimerson_unhelpful_2023}. Some studies advocate experimenting with multiple toolkits and hyperparameter configurations to identify solutions best suited to the language at hand. Indeed, while fully fine-tuning massive models can be effective, it often requires large amounts of computational resources, can risk overfitting with very small datasets, and demands updating millions of parameters.

Instead of modifying all model parameters, adapters introduce small trainable layers while keeping the base pre-trained model frozen, thereby reducing memory requirements and improving efficiency \citep{houlsby_parameter-efficient_2019}. This makes them particularly useful for linguistic fieldwork applications, where data is scarce and computational resources are limited. The MMS model developed by Meta \citep{pratap_scaling_2024} integrates adapter layers specifically designed for ASR, enabling efficient adaptation to new languages with minimal training data. Studies in low-resource settings \citep{bai_efficient_2024, mainzinger_fine-tuning_2024} have shown that adapter-based fine-tuning can achieve performance comparable to full fine-tuning while requiring significantly fewer trainable parameters. By avoiding overfitting on small datasets and focusing on the most relevant parameters for language adaptation, adapter-based methods offer an attractive balance between accuracy and efficiency, an approach increasingly vital to sustaining language documentation efforts in the face of extremely sparse resources.

In contrast to earlier studies that often focus on a single language or on clean, scripted corpora, our work systematically evaluates both MMS and XLS-R in truly low-resource fieldwork conditions spanning multiple typologically diverse languages. By examining noise-heavy, spontaneous recordings rather than controlled speech, we test model adaptability in settings that more accurately reflect real-world linguistic documentation. Further, our fine-grained error analysis explores how each model handles the nuanced phonological features that typify endangered and under-documented languages—details that have often been overlooked in prior research. Together, these innovations provide a clearer roadmap for linguists seeking practical ASR solutions under extreme data scarcity and diverse orthographic conventions.

\section{Data}
We test the performance of fine-tuned ASR models on five typologically varied low-resource languages: Cicipu (ISO639-3: awc, \citet{mcgill_cicipu_2012}), Mocho' (ISO639-3: mhc, \citet{perez_gonzalez_documentation_2018}), Toratán (ISO639-3: rth, \citep{jukes_documentation_2010}), Ulwa  (ISO639-3: yla, \citep{barlow_documentation_2018}), and Upper Napo Kichwa (ISO639-3: quw, \citep{grzech_upper_2020}). These languages span multiple language families and exhibit distinct phonetic, phonological, and morphological features. The data is drawn from the Endangered Languages Archive (ELAR)\footnote{\url{https://www.elararchive.org/}}, where gold-standard transcriptions can be derived from the recordings and the corresponding time aligned transcriptions in the ELAN \citep{brugman_annotating_2004} format. The dataset encompasses a variety of genres, such as greetings, narratives, ritual discourse, interviews, elicitation sessions, folktales, and cultural practices, the details of which are given in Table \ref{tab:language_genre} of Appendix \ref{sec:dataset_details}. Table \ref{tab: lang_detail} provides an overview of key linguistic features, including vowel and consonant inventories as well as tonal systems.

\begin{table*}[t]
  \centering
  \begin{tabular}{lllllllllll}
    \hline
    \textbf{Language} & \textbf{Family} & \textbf{Region} & \textbf{\#V} & \textbf{\#C} & \textbf{\#T} & \textbf{Features} \\
    \hline
    Cicipu  & Niger-Congo    & Nigeria         & 28  & 27  & 4   & \~{V}, V\textlengthmark, C\textlengthmark \\
    Mocho'   & Mayan          & Mexico          & 10  & 27  & 2   & V\textlengthmark \\
    Toratán  & Austronesian   & Indonesia       & 5   & 21  & 0   &  \\
    Ulwa     & Keram          & Papua New Guinea & 8   & 13  & 0  & [-voice, +son] \\
    Upper Napo Kichwa & Quechuan   & Ecuador        & 8   & 20  & 0  & V\textlengthmark \\
    \hline
  \end{tabular}
  \caption{\label{tab: lang_detail} Linguistic data used for the study, showing language family, region spoken, phoneme inventory size, tones, and phonological features such as nasality, vowel length, consonant gemination, etc.}
\end{table*}

\subsection{Data details}
Given that the recordings were made in naturalistic fieldwork environments, they exhibit acoustic idiosyncrasies that could pose significant challenges for ASR. There's background noise from outdoor settings, such as wind, animals, and community sounds, in many of the recordings. We also observe code-switching, usually in the regional dominant languages. For example, Toratán speakers are also speakers of Manado Malay (with various degrees of fluency), and loans from Malay are generally not adapted to Toratán phonology \citep{himmelmann_toratan_1999}. We also observe significant Spanish code-mixing in Mocho', as well as some English content in Cicipu.

The dataset sizes vary, with archived speech ranging from approximately 2 to 22 hours per language. However, not all archived data have been transcribed. This variation reflects real-world constraints in linguistic fieldwork, where some languages have more extensive documentation than others.

\subsection{Dataset Pre-processing}
All recordings were resampled to 16 kHz (the training sampling rate for both MMS and XLS-R), converted to mono-channel WAV format, and aligned with their corresponding transcriptions. During transcription pre-processing, we referenced the phonological description of each language to ensure that punctuation marks or special characters used to denote phonological features were retained (see Table \ref{tab:language_genre} of Appendix \ref{sec:dataset_details}). Audio segments explicitly transcribed as non-linguistic sounds, such as laughter, were excluded from the dataset. We also removed utterances that contain only filler words, such as 'mhm', 'aaa', etc. A breakdown of the audio lengths before and after pre-processing is given in Table \ref{tab:data_audio_length} of Appendix \ref{sec:dataset_details}

We created four total train+dev duration configurations for each language—10, 30, 60, and 120 minutes—before splitting the data into training (90\%) and development (10\%) sets. In addition, we set aside a fixed 10-minute test set for final evaluation. To maintain consistency and facilitate interpretation, larger dataset splits were structured as supersets of smaller ones. We did not designate a held-out speaker, as field linguists typically work with a limited number of consultants and would prioritize consistent model performance across familiar speakers \citep{liu_investigating_2023}. Details of the cleaned dataset are shown in Table \ref{tab:processed_data}. The last column of the table lists the number of unique characters used in the language. Due to different transcription conventions, features such as nasalization, vowel length and voicing could be indicated with diacritics or extra letters. Therefore, although transcriptions are meant to be phonemic, the number of unique characters might not match the number of contrastive vowels and consonants. In the case of Cicipu, its unusually large inventory of 93 characters is due to the number of all possible combinations of nasality, tone, and vowel quality marking.

\begin{table}[t]
  \centering
  \begin{tabular}{lcccc}
    \hline
    \textbf{Language} & \textbf{\#Spk} & \textbf{Avg. Leng.} & \textbf{\#Char} \\
    \hline
    Cicipu              & 33  & 2.1  & 93  \\
    Mocho'             & 6   & 2.0  & 29  \\
    Toratán            & 13  & 2.35 & 27  \\
    Ulwa               & 6   & 3.65 & 25  \\
    U.N. Kichwa  & 16  & 3.79 & 33  \\
    \hline
  \end{tabular}
  \caption{\label{tab:processed_data} Dataset statistics for different languages, including the number of speakers, average utterance length in seconds, and character inventory.}
\end{table}

\section{Methodology}
This study investigates the effectiveness of fine-tuning multilingual ASR models to address the unique challenges posed by low-resource linguistic fieldwork recordings. By evaluating the performance of two state-of-the-art models, we aim to determine how fine-tuning can enhance recognition accuracy on typologically diverse, low-resource languages. In addition, we discuss the impact of key factors, including training data size, model choice, and pre-trained model features, to provide practical insights for ASR adaptation in fieldwork contexts.

\subsection{Models}
Our goal is to fine-tune state-of-the-art multilingual ASR models that have been pre-trained on large-scale speech corpora. Specifically, we evaluate models from Meta's Massively Multilingual Speech (MMS) project \citep{pratap_scaling_2024} alongside XLS-R \citep{babu_xls-r_2021}, a widely used multilingual ASR model. 

MMS is based on the wav2vec 2.0 framework \citep{baevski_wav2vec_2020}, which employs self-supervised learning to extract generalized speech representations from vast amounts of unlabeled audio. The model has been trained on 1,406 languages, making it one of the most comprehensive ASR models for multilingual speech recognition. Specifically, we choose MMS-1B-l1107, a 1-billion parameter model fine-tuned specifically for ASR with an additional 2-million parameter adapter, which supports ASR in 1107 languages out of the box \citep{houlsby_parameter-efficient_2019}. The adapter facilitates efficient language-specific fine-tuning while preserving the generalized multilingual knowledge encoded in the base model.

XLS-R \citep{babu_xls-r_2021} is a multilingual ASR model pre-trained on 128 languages using the wav2vec 2.0 framework. It has been extensively used for low-resource ASR and cross-lingual transfer learning, making it a strong baseline for evaluating ASR performance in linguistic fieldwork settings. Unlike MMS, which is trained on over 1,000 languages, XLS-R has been optimized for a balanced selection of 128 languages, with a strong focus on phonetic diversity. This makes it particularly useful for comparison against MMS models to assess the effectiveness of scaling multilingual pre-training to extremely low-resource languages. The XLS-R-300m with 300 million parameters is chosen for the study. 

None of the languages used in the study, with the exception of Upper Napo Kichwa, is represented in the training data of the two models. A discussion of the possible effects is included in Section \ref{subsec:mms_vs_xlsr}. 

\subsection{Implementation}
Following the fine-tuning procedure outlined by von Platen\footnote{\url{https://huggingface.co/blog/mms_adapters}} and using the Hugging Face Transformers library \citep{wolf_transformers_2020}, we implement model-specific strategies. For MMS-1B-l1107, only the adapter layers are fine-tuned, with the base model frozen. For XLS-R, the entire model is fine-tuned. To assess the effect of data size, models are trained on subsets of 10, 30, 60, and 120 minutes of transcribed fieldwork data per language. Early stopping, based on the development set Character Error Rate (CER), mitigates overfitting. Hyperparameter details, including batch size, learning rate, optimization strategy, and training details are provided in Appendix \ref{sec:appendix}.

\section{Results}

\begin{figure}[t]
  \includegraphics[width=\columnwidth]{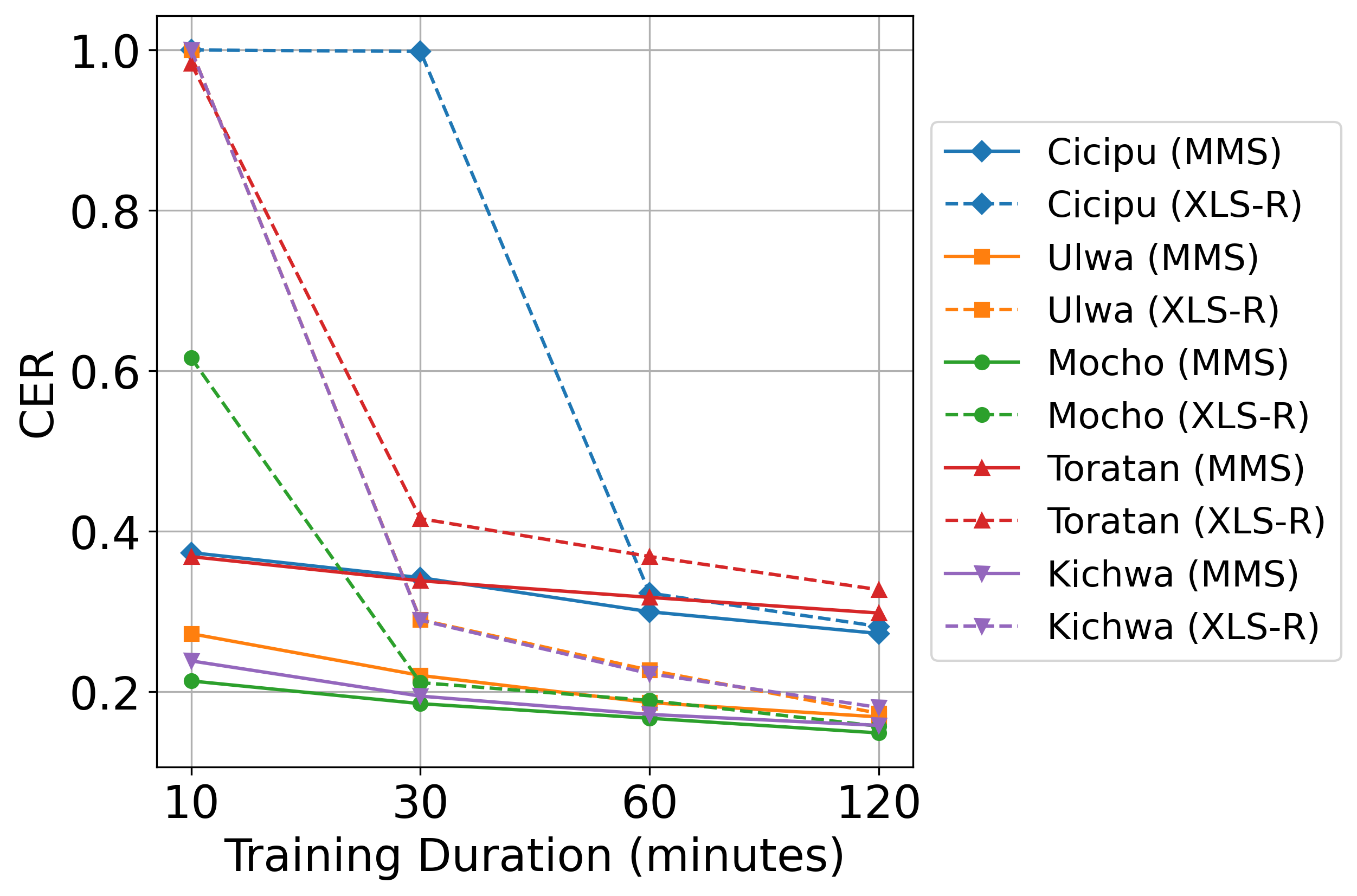}
  \caption{CER comparison for MMS-1b1107 and XLS-R-300m models across five languages. The MMS model performs markedly better under extremely low-resource settings (less than 1 hour), but XLS-R performs similarly well with 2 hours of data. Points are connected to aid trend reading and do not imply performance at intermediate durations.}
  \label{fig:cer_comparison}
\end{figure}

\begin{figure}[t]
  \includegraphics[width=\columnwidth]{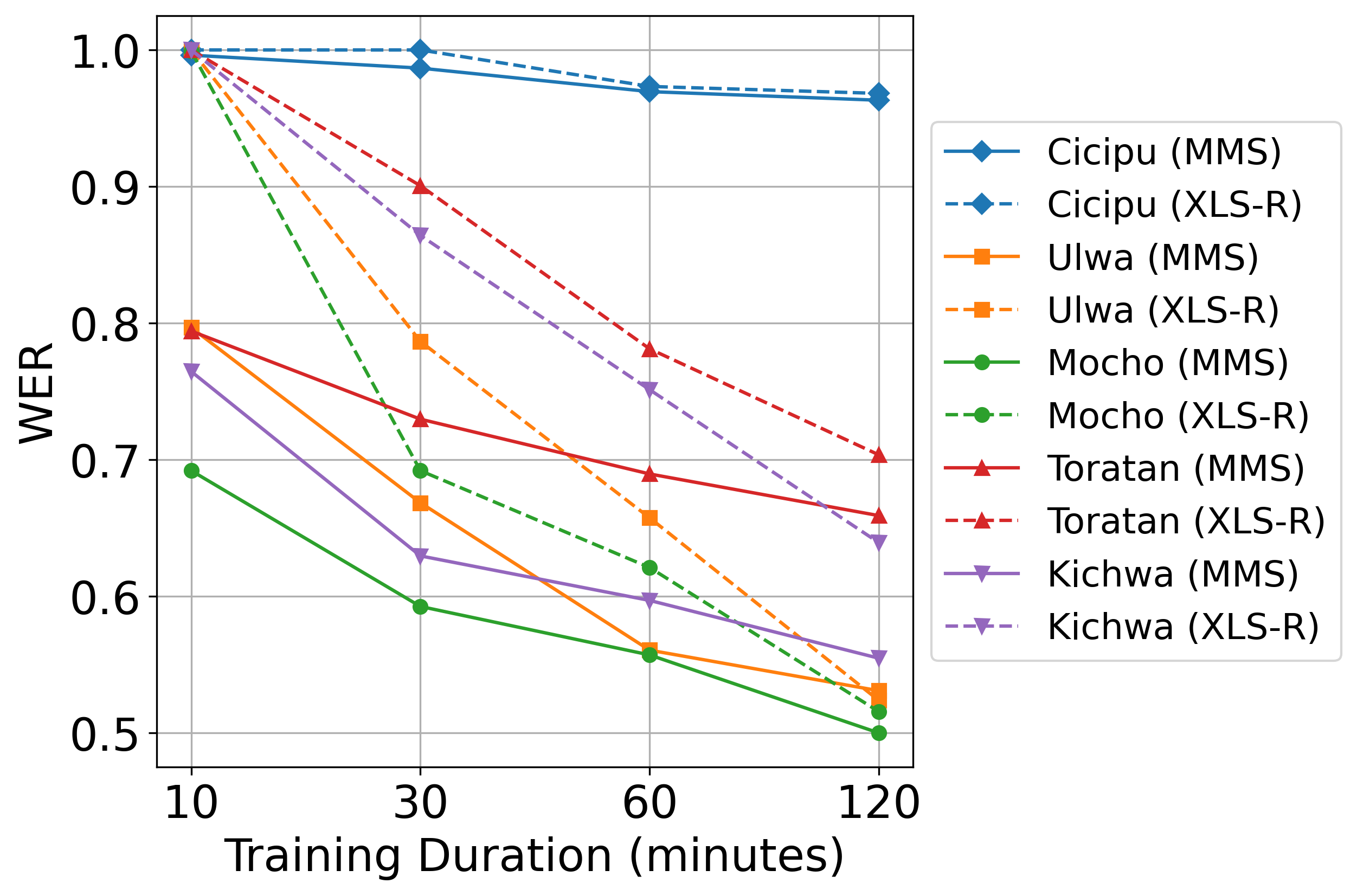}
  \caption{WER comparison for MMS-1b1107 and XLS-R-300m models across five languages.}
  \label{fig:wer_comparison}
\end{figure}

Overall, the performance of the MMS and XLS-R models in automatic speech recognition (ASR) tasks on low-resource language data is comparable, though nuanced differences emerge depending on the availability of training data. In general, the MMS model outperforms XLS-R under extremely low-resource conditions (i.e., less than one hour of transcribed data). However, XLS-R demonstrates a marked improvement as the size of the training data scales beyond this threshold, ultimately becoming on par with MMS. Figure \ref{fig:cer_comparison} and Figure \ref{fig:wer_comparison} illustrate these trends in Character Error Rate (CER) and Word Error Rate (WER) metrics across the five languages in this study.

In our analysis, CER serves as the primary evaluation metric. Unlike WER, which operates at the word level and often presupposes well-defined word boundaries and stable orthographic forms, CER better reflects the needs of field linguists, who often lack enough data to be able to use language models. In fieldwork contexts, the primary goal of transcription is to capture phonetic accuracy, particularly in languages without standardized orthographies. During model training, the best model metric is set to CER to ensure that model performance aligns with the core task of producing reliable phoneme-level transcriptions from spontaneous and noisy speech data. 

\subsection{Data size effect}
The relationship between training data size and model performance is critical in understanding model suitability for field linguistics applications. For both MMS and XLS-R, performance improvements start to plateau when training data exceeds approximately one hour. This finding suggests steady although diminishing returns beyond this point, aligning with previous observations in low-resource ASR research \citep{guillaume_plugging_2022}. 

In scenarios where less than one hour of data is available, MMS consistently achieves lower error rates, likely due to its extensive multilingual pre-training on over 1,000 languages. For field linguists dealing with extremely limited resources, we thus recommend fine-tuning MMS to achieve acceptable ASR performance. However, when approximately one hour of data or more is obtainable, XLS-R becomes a more effective option due to its improving performance with increasing data volumes. This suggests that one hour of transcribed data serves as a practical threshold for developing a robust fine-tuned ASR system in fieldwork contexts. 

It is worth noting that Cicipu exhibits particularly high error rates under extreme data scarcity, including a character error rate approaching 1.0 at 10 minutes of data for both models and at 30 minutes for XLS-R. Cicipu's unusually large orthographic inventory (93 unique characters reflecting combinations of nasality, tone, and vowel quality) requires more training examples to accurately learn the mapping from acoustics to graphemes. Consequently, with only a few minutes of labeled data, neither model can fully learn Cicipu's complex phonological and orthographical features.

\subsection{MMS vs.\ XLS-R}
\label{subsec:mms_vs_xlsr}
The MMS model excels in settings with minimal data due to its multilingual pretraining on a vast corpus that includes low-resource languages. Moreover, unlike XLS-R—whose core version is primarily self-supervised and not initially fine-tuned for ASR—MMS has already undergone a large-scale ASR fine-tuning step. This means that MMS starts off with more task-specific parameters, making it more effective than XLS-R in extremely low-data regimes. However, MMS’s reliance on primarily read speech data (e.g.\ Bible translations) may limit its adaptability to spontaneous speech environments, which are common in linguistic fieldwork recordings.

In contrast, XLS-R benefits from a more diverse training corpus that encompasses conversational and spontaneous speech, allowing it to generalize better once sufficient data becomes available. Indeed, \citet{mainzinger_fine-tuning_2024} reported superior performance of MMS over XLS-R when fine-tuning Mvskoke—likely due to both the advantage of MMS’s ASR fine-tuning and the fact that much of the Mvskoke training material was similarly read or scripted speech.

Since several related dialects of Kichwa \citep{eberhard_napo_2024, eberhard_tena_2025} were included in the MMS pre-training dataset, we investigated whether the performance gap between MMS and XLS-R would be \emph{larger} for Kichwa than for the other languages in our study. Specifically, we fit a linear mixed-effects model (with random intercepts for each language) to our character error rate (CER) data, using \textit{model} (MMS vs.\ XLS-R), \textit{time}, and an indicator \textit{similar} (1 = Kichwa, 0 = other languages) as fixed effects. If Kichwa had benefitted disproportionately from MMS’s pre-training, we would have observed a significant positive interaction in the model. However, the interaction term (\textit{model}~$\times$~\textit{similar}) was small and not statistically significant ($\beta = 0.021,\, p=0.896$), indicating that while MMS outperforms XLS-R overall, the additional advantage for Upper Napo Kichwa is not discernibly greater than for the other languages.

Further research is needed to evaluate whether the performance trend continues with larger datasets, particularly for languages with similar phonological and morphological complexity as those in this study. Additionally, the effectiveness of adapter-based fine-tuning for MMS suggests that optimizing model architecture for scalable adaptation could yield further improvements.

\subsection{Error analysis}

We performed a phonologically informed error analysis on two of the languages in our dataset, Cicipu and Mocho', both of which exhibit segmental contrasts that could be challenging for ASR models. Cicipu's orthography explicitly marks tone and nasality with diacritics and differentiates both vowel and consonant length with doubled letters (e.g., 'aa' for long vowels, 'tt' for geminate consonants), making it well suited for evaluating the system's performance on these phonological categories. Mocho' similarly features a vowel length distinction encoded with doubled vowel letters. Other languages in our dataset, such as Ulwa and Upper Napo Kichwa, contain too few instances of long vowels or voiceless sonorants for a robust category-based analysis.

\subsubsection{Error rates}
To quantify performance on these features, we leverage character-level alignments (details in Appendix~\ref{sec:error_rates_formulas}) and calculate phonological segment error rates (Table~\ref{tab:phonological_errors}). For each category \(C \in \{\text{Tone}, \text{Nasality}, \text{V\_length}, \text{C\_length}\}\), we sum all substitutions, deletions, and insertions that affect that category and normalize by the total number of reference tokens \(L_C\) exhibiting \(C\).

\begin{table}[t]
\centering
\small
\setlength{\tabcolsep}{2pt}
\begin{tabular*}{0.88\columnwidth}{@{\extracolsep{\fill}} l l c c c c @{}}
\toprule
\textbf{Lang} & \textbf{Model} & \textbf{Tone} & \textbf{Nas.} 
              & \textbf{V-Len} & \textbf{C-Len} \\ 
\midrule
    Cicipu & MMS   & 0.199 & 0.337 & 0.239 & 0.174 \\
           & XLS-R & 0.215 & 0.337 & 0.248 & 0.183 \\
\midrule
    Mocho' & MMS   & ---   & ---   & 0.154 & ---   \\
           & XLS-R & ---   & ---   & 0.160 & ---   \\
\bottomrule
\end{tabular*}
\caption{Phonological error rates (0--1 scale) for Cicipu and Mocho'. 
Cicipu shows higher confusion in tone, nasality, and consonant length, 
while Mocho' displays more issues with vowel length. Both models 
(MMS vs.\ XLS-R) yield broadly similar error patterns.}
\label{tab:phonological_errors}
\end{table}

As shown in Table~\ref{tab:phonological_errors}, both MMS and XLS-R struggle with Cicipu’s tone, nasality, and consonant length, each exhibiting error rates in the range of 30--38\%. By contrast, vowel-length confusion is comparatively low (7--9\%). For Mocho', the long--short vowel distinction remains problematic, with error rates around 35--38\%. These findings suggest that neither model has a strong advantage for these particular phonological categories; even after fine-tuning, nuanced contrasts such as nasality and tone remain challenging.

\subsubsection{Error distribution}
For further investigation, we performed a more detailed error analysis on the output from the 120-minute model of XLS-R for Cicipu and Mocho'. Table~\ref{tab:error_distribution} breaks down each category by the percentage of substitutions (\textit{S}), deletions (\textit{D}), and insertions (\textit{I}).

\begin{table}[t]
\centering
\small
\setlength{\tabcolsep}{5pt}
\begin{tabular}{l l c c c}
\toprule
\textbf{Lang} & \textbf{Category} & \textbf{S\%} & \textbf{D\%} & \textbf{I\%} \\
\midrule
Cicipu & Tone       & 47.62 & 32.11 & 20.27 \\
 & Nasality   &  0.00 & 70.00 & 30.00 \\
 & V-Len  & 55.58 & 27.86 & 16.56 \\
 & C-Len   & 44.18 & 33.90 & 21.92 \\
\midrule
Mocho' & V-Len  & 58.29 & 26.86 & 14.86 \\
\bottomrule
\end{tabular}
\caption{Percentage of substitutions (S\%), deletions (D\%), and insertions (I\%) 
in XLS-R 120\,min output, for each phonological category in Cicipu and Mocho'.}
\label{tab:error_distribution}
\end{table}

In Cicipu, nearly half of the tone errors (\(\sim48\%\)) are substitutions, indicating confusion over which tone mark to apply, while around a third are deletions and the remainder insertions. Nasality errors, by contrast, skew heavily toward deletions (\(\sim70\%\)), suggesting the model often fails to detect nasal vowel features. Substitutions are rare for nasality, indicating that the system either omits it entirely or adds it spuriously rather than confusing it with another tone diacritic. Vowel-length errors (\(\sim44\%\) for deletions and insertions) and consonant-length errors (\(\sim55\%\) for deletions and insertions) reflect a high level of segment-level confusion, whereas confusion with a different vowel (\(\sim56\%\) substitutions) or consonant (\(\sim44\%\) substitutions) is also frequent. For Mocho’, vowel-length confusion is likewise dominated by substitutions (\(\sim58\%\)), a pattern similar to Cicipu which reveals that XLS-R often misidentifies one segment in the long vowel. The distributions point out that while the model does capture some acoustic correlates of nasality, tone, and length, it nevertheless struggles to map them consistently to diacritics and extended graphemes in low-resource scenarios.

Overall, these patterns highlight the persistent challenge of representing languages with complex orthographies and rich phonological inventories. Even after multilingual pre-training and fine-tuning, contrasts such as tone or nasality may be overlooked when the amount of transcribed data is minimal. Addressing these gaps may require linguistically informed data augmentation, specialized adapter modules, or loss functions that explicitly emphasize distinct phonological categories. In extremely low-resource settings, such targeted methods could provide the additional examples and acoustic cues needed for more accurate transcription of endangered languages.

\section{Conclusion} 

Our experiments show that fine-tuned multilingual ASR models can substantially reduce the transcription burden for endangered and low-resource languages. Across five typologically diverse languages, MMS proved more effective with extremely limited labeled data, whereas XLS-R caught up once approximately one hour of transcribed material was available. By using Character Error Rate (CER) rather than Word Error Rate (WER), we focus on phoneme-level accuracy—a more direct measure for languages without standardized orthographies. Despite improvements in overall accuracy, both models struggled with challenging phonological categories in Cicipu, such as tone and consonant length, and exhibited a high rate of vowel-length confusion in Mocho'. These findings confirm that current multilingual ASR systems are indeed helpful for language documentation but still require targeted adaptations to handle nuanced phonological contrasts in under-resourced settings.

\section{Future Work}

Although our study confirms that fine-tuning multilingual ASR models can substantially reduce transcription overhead for low-resource languages, several research directions remain promising for further performance gains. One compelling approach is continued pre-training (CoPT) on unlabeled in-language audio. \citet{dehaven_improving_2022} show that CoPT on a wav2vec 2.0--based multilingual model can match or outperform pseudo-labeling techniques while being more computationally efficient. Similarly, \citet{nowakowski_adapting_2023} demonstrate that CoPT on about 234 hours of Sakhalin Ainu audio yields a considerable reduction in error, beyond what standard multilingual fine-tuning achieves. CoPT has also proven effective in domain adaptation, especially for noisy data or new speaker types \citep{attia_continued_2024}. While the scarcity of fieldwork data could limit the scale of CoPT, even incremental benefits may substantially ease the manual transcription effort.

A second avenue is leveraging diverse augmentation methods to enlarge the effective training set. Self-training (pseudo-labeling) uses an initial ASR model to generate transcripts for unlabeled audio, which can then be added to the training pool. This method has consistently boosted low-resource ASR performance \citep{bartelds_making_2023}, particularly when coupled with filtering or iterative refinement. TTS-based augmentation offers another option: if a target-language text-to-speech system is available, synthesizing speech from text yields additional “perfectly labeled” data, potentially improving recognition robustness (ibid). Finally, common audio perturbations, speed/pitch changes, SpecAugment, and noise injection, remain valuable for avoiding overfitting and preparing the model for real-world variability.

A final challenge involves better capturing difficult phonological categories, such as tone, nasality, and consonant length. Adapters in MMS could be extended or reconfigured to emphasize language-specific features, while training regimes could incorporate acoustic or phonological priors explicitly. Future work might integrate fine-grained linguistic annotations (if available) or employ specialized masking strategies during CoPT to boost the model’s sensitivity to subtle contrasts. Combining these techniques into user-friendly toolkits will be essential for widespread adoption by field linguists, who often have limited computational resources yet require high-accuracy, phoneme-level transcriptions for documenting and revitalizing endangered languages.

\section{Limitations}
Despite promising results, several specific limitations affect the generalizability and applicability of our approach. The most critical limitation is related to the data size and representativeness of the linguistic diversity considered. Our study focused on a small number of typologically diverse languages, each with relatively limited datasets ranging from just a few minutes to two hours. As such, the models’ performances may not generalize to other endangered or low-resource languages with distinct phonological or orthographic features.

Additionally, due to constraints inherent in linguistic fieldwork, the training and test datasets often contained data from the same speakers, potentially inflating model accuracy estimates. Future research should validate these findings with genuinely held-out speakers to better gauge model robustness to speaker variability.

Moreover, the orthographic inconsistencies and the absence of standardized orthographies in our datasets likely influenced model performance, especially for phonologically complex categories like tone, vowel length, and nasality. This issue highlights a broader limitation: ASR models trained under these conditions may struggle to generalize to spontaneous and noisy field recordings, especially when orthographic conventions vary within and across datasets. 

Finally, computational resource limitations (training on a single NVIDIA T4 GPU with constrained runtimes) restrict our ability to fine-tune larger models or extensively optimize hyperparameters, which may have further improved performance. Addressing these limitations would require additional computational resources and potentially more extensive data augmentation strategies tailored explicitly to low-resource linguistic contexts.

\bibliography{custom}

\appendix

\section{Training Details}
\label{sec:appendix}
The models were trained on an NVIDIA T4 GPU, with training times ranging from approximately 1 to 60 minutes per model. The hyper-parameters were defined as follows:
\begin{itemize}
    \item \textbf{Learning rate:} MMS: 1e-3; XLS-R: 3e-4
    \item \textbf{Maximum epochs:} 30
    \item \textbf{Best model metric:} Character Error Rate (CER)
    \item \textbf{Early stopping:} 3 epochs
    \item \textbf{Early stopping threshold:} 0.003
\end{itemize}

\section{Dataset Details}
\label{sec:dataset_details}

Table \ref{tab:language_genre} provides detailed information on the genres and types of content present in the datasets used in this study, along with key linguistic references and citations to the original documentation archives. Table \ref{tab:data_audio_length} summarizes the total archived hours of audio recordings for each language and the amount of data remaining after the cleaning and preprocessing steps described earlier in the manuscript.

\begin{table*}[t]
  \centering
  \begin{tabular}{lp{4cm}p{4cm}p{5cm}}
    \hline
    Language & Genres of content & Phonological Description & Documentation \\
    \hline
    Cicipu & 
    Greetings, conversations, hortative discourse, narratives, procedural discourse, ritual discourse, elicitation activities & 
    \citet{mcgill_cicipu_2014} & \citet{mcgill_cicipu_2012} \\
    \hline
    Mocho' & 
    Biographical and non-biographical narratives (historical events, myths, local beliefs, traditional building, witchcraft), prayer, conversation, elicitation sessions, text translation & 
    \citet{palosaari_topics_2011} & \citet{perez_gonzalez_documentation_2018} \\
    \hline
    Toratán & 
    Conversational data, elicitation sessions, narratives (personal history, folk tales) & 
    \citet{himmelmann_toratan_1999} & \citet{jukes_documentation_2010} \\
    \hline
    Ulwa & 
    Conversational data, traditional stories, personal stories, traditional singing and dancing video & 
    \citet{barlow_grammar_2018} & \citet{barlow_documentation_2018} \\
    \hline
    Upper Napo Kichwa & 
    Grammatical elicitation, life interviews & 
    \citet{wroblewski_amazonian_2012}, \citet{orourke_tena_2013} & \citet{grzech_upper_2020} \\
    \hline
  \end{tabular}
  \caption{Details of depository content for languages used in this paper, related linguistic work referenced, and original documentation citations.}
  \label{tab:language_genre}
\end{table*}

\begin{table}[h]
    \centering
    \begin{tabular}{lcc}
        \hline
        \textbf{Language} & \textbf{Total Hours} & \textbf{Cleaned Hours} \\
        \hline
        Cicipu & 5.66 & 3.09 \\
        Mocho' & 7.26 & 4.21 \\
        Toratán & 22.84 & 11.15 \\
        Ulwa & 3.25 & 2.83 \\
        Upper Napo Kichwa & 13.19 & 6.97 \\
        \hline
    \end{tabular}
    \caption{Total archived and cleaned hours of audio for all languages used in the study.}
    \label{tab:data_audio_length}
\end{table}

\section{Error Rates}
\label{sec:error_rates_formulas}
We consider four phonological categories:
\[
  C \in \{\text{Tone},\,\text{Nasality},\,\text{V\_length},\,\text{C\_length}\}.
\]
Over the entire dataset, we record:
\begin{itemize}
  \item $S_C$: the total \emph{substitution} errors for category $C$,
  \item $D_C$: the total \emph{deletion} errors for category $C$,
  \item $I_C$: the total \emph{insertion} errors for category $C$,
  \item $L_C$: the total \emph{reference tokens} exhibiting category $C$ 
    (e.g.\ \texttt{tone\_labels} for tone, \texttt{total\_vowels} for vowel length, etc.).
\end{itemize}

We then define the total errors and error rate for each category $C$ as follows:
\[
  E_C \;=\; S_C \;+\; D_C \;+\; I_C
  \quad\text{and}\quad
  \text{ErrorRate}_C \;=\; \frac{E_C}{L_C}.
\]

For example, if $C=\text{tone}$ then $L_C=\text{tone\_labels}$ (the number of reference tokens with at least one tone diacritic). Similarly, if $C=\text{vowel\_length}$ then $L_C=\text{total\_vowels}$ (the total vowel tokens in the reference).

\end{document}